\documentclass[runningheads]{llncs}
\usepackage[T1]{fontenc}
\usepackage{graphicx,verbatim}
\usepackage{multirow}
\usepackage{booktabs}
\usepackage{url}
\usepackage{cite}
\usepackage{algpseudocode}
\usepackage{tabto}
\usepackage{amsmath} 
\usepackage{algorithm}
\usepackage{pifont} 
\usepackage[hidelinks]{hyperref}
\usepackage{adjustbox} 
\usepackage{booktabs}
\usepackage{arydshln} 
\usepackage{lmodern}

\begin{document}
\title{X2CT-CLIP: Enable Multi-Abnormality Detection in Computed Tomography from Chest Radiography via Tri-Modal Contrastive Learning}
%

\author{Jianzhong You$^{1,2,5,7}$, Yuan Gao$^{1,2,4,5,7}$, Sangwook Kim$^{1,2,5,7}$, Chris Mcintosh$^{1,2,3,4,5,6,7}$}  
\institute{$^1$ Peter Munk Cardiac Centre, University Health Network (UHN), Toronto, Canada \\
           $^2$ Department of Medical Biophysics, University of Toronto (U of T), Toronto, Canada \\
           $^3$ Department of Computer Science, U of T, Toronto, Canada \\
           $^4$ Ted Rogers Centre for Heart Research, UHN, Toronto, Canada \\
           $^5$ Toronto General Hospital Research Institute, UHN, Toronto, Canada \\
           $^6$ Department of Medical Imaging, U of T, Toronto, Canada \\
           $^7$ Vector Institute, Toronto, Canada \\
    \email{\{Jianzhong.you, yuan.gao, sangwook.kim, chris.mcintosh\}@uhn.ca}
}

\maketitle              
\begin{abstract}
Computed tomography (CT) is a key imaging modality for diagnosis, yet its clinical utility is marred by high radiation exposure and long turnaround times, restricting its use for larger-scale screening. Although chest radiography (CXR) is more accessible and safer, existing CXR foundation models focus primarily on detecting diseases that are readily visible on the CXR. Recently, works have explored training disease classification models on simulated CXRs, but they remain limited to recognizing a single disease type from CT. CT foundation models have also emerged with significantly improved detection of pathologies in CT. However, the generalized application of CT-derived labels on CXR has remained illusive. In this study, we propose X2CT-CLIP, a tri-modal knowledge transfer learning framework that bridges the modality gap between CT and CXR while reducing the computational burden of model training. Our approach is the first work to enable multi-abnormality classification in CT, using CXR, by transferring knowledge from 3D CT volumes and associated radiology reports to a CXR encoder via a carefully designed tri-modal alignment mechanism in latent space. Extensive evaluations on three multi-label CT datasets demonstrate that our method outperforms state-of-the-art baselines in cross-modal retrieval, few-shot adaptation, and external validation. These results highlight the potential of CXR, enriched with knowledge derived from CT, as a viable efficient alternative for disease detection in resource-limited settings.
\end{abstract}

\def\thickhline{\noalign{\hrule height0.75pt}}
\section{Introduction}
Medical imaging is crucial in diagnosing and managing various diseases, including cardiovascular conditions, lung pathologies, and many cancers. While computed tomography (CT) is a powerful tool for disease detection and risk assessment, it has notable drawbacks that limit its applicability in routine screening, including longer turnaround times for image acquisition and interpretation, and higher costs and dosages of ionizing radiation, which can pose health risks. In contrast, chest radiography (CXR) is more widely accessible and cost-effective. In particular, it emits significantly lower radiation, making it a safer and more practical alternative for patients in many clinical settings. Given these clinical advantages, this study explores the feasibility of leveraging solely CXR to predict diseases that are traditionally only identifiable in CT, aiming to reduce reliance on CT while enabling earlier detection, improving patient outcomes, and optimizing healthcare resources.

The development of Contrastive Language-Image Pretraining (CLIP)~\cite{clip} demonstrated the effectiveness of contrastive learning (CL) on large-scale image-text pairs, allowing robust generalization in diverse downstream tasks. The success of CLIP has led to the development of several works in the CXR domain, including GLoRIA \cite{gloria}, MedCLIP \cite{medclip}, and CXR-CLIP \cite{cxrclip}. All of these involve the alignment of CXR and clinical text knowledge in latent space. CLIP also inspired research into multi-modal CL beyond two modalities in the medical imaging domain. MEDBind \cite{medbind} introduced tri-modal contrastive learning to unify CXR, electrocardiograms, and text, enhancing cross-modal binding with its Edge-Modality Contrastive Loss. Building on the success of CL in 2D medical imaging, recent advances have extended these techniques to develop foundation models in 3D CT. These models, such as FM-CT \cite{fm-ct} and CT-CLIP \cite{ctclip}, leveraged large-scale text-paired CT datasets and CL to develop generalizable embeddings that enabled multi-abnormality classification on CT. These models underscore the versatility of multi-modal CL across various domains.

Two key limitations remain in classification models for CT-level disease. \textbf{First}, although CT foundation models exist, their utilization necessitates the acquisition of CT images, suffering from the aforementioned acquisition and radiation drawbacks. \textbf{Second}, although foundation models for CXRs have been extensively studied to predict a wide range of CXR-diagnosed diseases, no model has attempted to predict multiple CT-diagnosed conditions from CXR. Closely related works in this regard include \cite{synXrayLungDisease} and BI-Mamba \cite{bimamba}. Specifically, \cite{synXrayLungDisease} leverages simulated CXRs to enhance model performance in lung cancer classification, while \cite{bimamba} employs a state-space model \cite{mamba} to predict cardiovascular disease (CVD) found in CT images from simulated CXRs. Unfortunately, both approaches are limited to a single type of CT pathology, underscoring the need for a foundation model capable of capturing diverse CT-level disease knowledge based solely on CXRs, thereby enabling the development of more scalable and effective screening tools in clinical settings.

\textbf{Contributions:} We propose X2CT-CLIP, the first CL framework that bridges the modality gap from chest radiography to CT (X2CT) to align CXR with their corresponding CT and CT report in latent space, enabling the detection of multiple traditionally CT-diagnosed abnormalities from CXR. Our approach leverages the features of CT reports and 3D CT volumes derived from CT-CLIP \cite{ctclip} to enrich a CXR encoder via cross-modal knowledge transfer, while reducing the demand for computational resources in model training. We evaluated X2CT-CLIP on three multi-label (multi-abnormality) CT datasets, including CT-RATE \cite{ctclip}, RadChest-CT \cite{radchest_ct}, and a curated MIMIC-CT dataset from MIMIC \cite{mimic-cxr}. Our results demonstrated that X2CT-CLIP outperforms state-of-the-art baselines in cross-modal retrieval, few-shot adaptation, and external validation, highlighting its potential to reduce reliance on CT imaging for regular disease screening and the broader potential for 2D-to-3D data alignment in medical imaging.

\begin{figure}[t!]
    \centering
    \includegraphics[width=\textwidth]{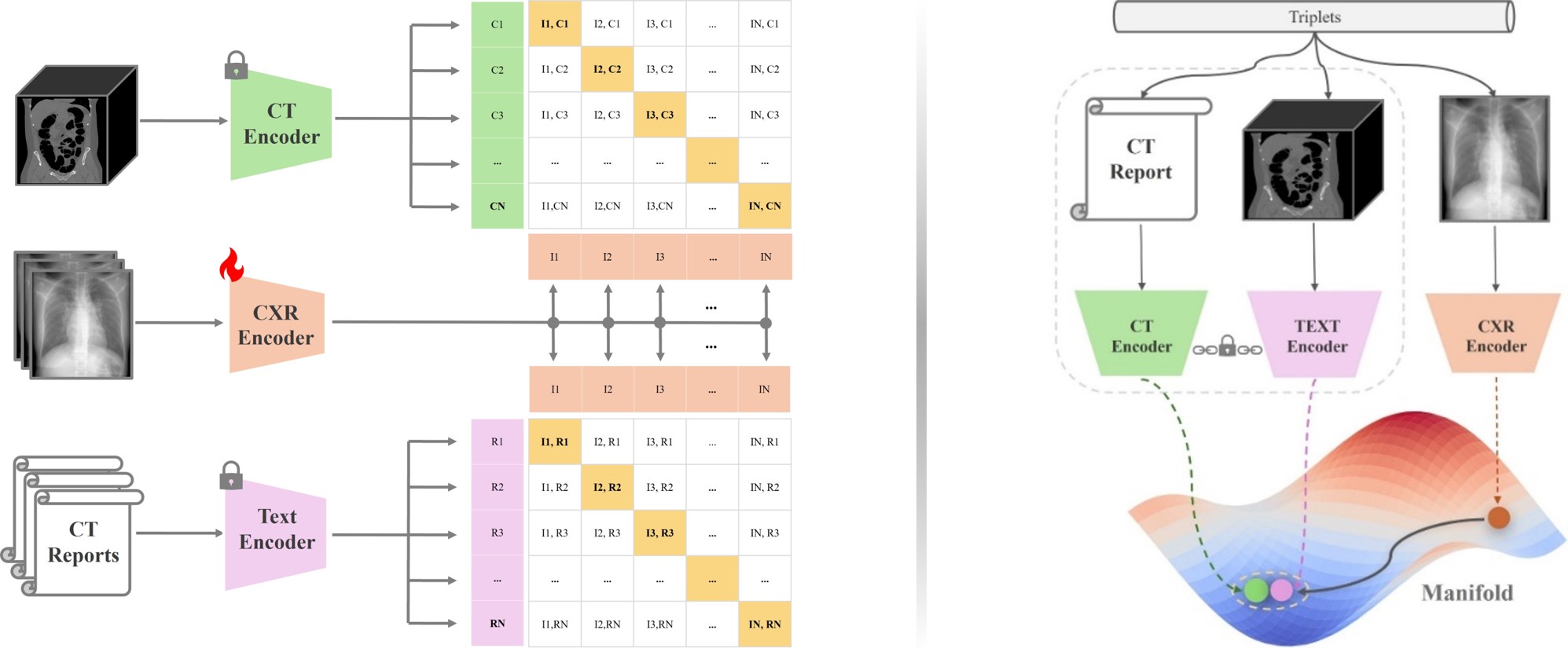}
    \caption{\textbf{Left:} Tri-modal contrastive learning framework of the CT, CT report, and CXR triplet. \textbf{Right:} Latent space alignment of CT, CT report, and CXR features.} 
    \label{fig:method}
\end{figure}
\section{Method}
This section first describes constructing CT, CT report, and CXR triplets for our X2CT-CLIP. We then provide a detailed description of our method for pretraining the CXR encoder in X2CT-CLIP while reducing hardware requirements.
\subsection{Creating CT, CT Report, and CXR Triplets} \label{sec:synxray}
Datasets with CXRs, CT images and label pairings are not publicly available. However, recent advancements in the 3D point cloud domain demonstrated that models operating in data-constrained settings can significantly enhance their recognition capabilities by incorporating knowledge from other modalities that share the same semantic meaning \cite{ulip, ulipv2}.

Motivated by this merit, we created simulated CXRs from real CTs in CT-RATE dataset~\cite{ctclip} for model training. A CT scan captures cross-sectional slices of the body that calculate Hounsfield units of radiation attenuation per voxel, making it possible to generate simulated CXR images from CT data with appropriate computational processing~\cite{bimamba, synXrayLungDisease}. We created our simulated CXR datasets to pair with known CTs and reports from CT-RATE~\cite{ctclip}. We take each CT from CT-RATE and simulate the corresponding anteroposterior view of CXR using \cite{simpleITK}. We generated $50188$ triplets $T_i=(C_i, R_i, X_i)$ of CTs, CT reports, and CXRs, respectively, to pretrain our CXR encoder in the X2CT-CLIP framework.
\subsection{Unifying Latent Space of CT, CT Report, and CXR}
By pretraining on the CT-RATE dataset, which includes CT-Report pairs covering 13 diseases, CT-CLIP learned generalizable semantic features in CT that demonstrated superior performance in downstream tasks. By leveraging CT-CLIP, we propose to align the feature representations of CT volumes and CT reports with CXRs as a unified latent space. As illustrated in Figure~\ref{fig:method}-Right, the CXR feature space is integrated into the pre-aligned representation space of CT and the CT report by freezing the weights of CT-CLIP, while fine-tuning the CXR encoder. This allows a seamless integration of CXR features while preserving the pretrained knowledge from CT-CLIP.

To achieve this, we adopt tri-modal contrastive learning to align feature embeddings from modality-specific encoders within a shared latent space. We take advantage of the pre-aligned CT encoder $f_{C}(\cdot)$ and CT report encoder $f_{R}(\cdot)$ from CT-CLIP (where $f_{C}(\cdot)$ is a 3D vision transformer, 3DViT \cite{ct-vit} and $f_{R}(\cdot)$ is a CXR-BERT \cite{cxr-bert}) and train a CXR encoder (ResNet~\cite{resnet} or Swin~\cite{swin}) $ f_{X}(\cdot) $ from scratch. Given a triplet $T_i=(C_i, R_i, X_i)$, the respective feature representations are obtained as $ h^{\text{C}}_i = f_{\text{C}}(C_i) $, $ h^{R}_i = f_{R}(R_i) $, and $ h^{X}_i = f_{X}(X_i) $. The model enforces feature similarity between pairs of modalities by optimizing the InfoNCE \cite{infoNCE} loss defined in Eq~\ref{eq:infoNCE}:
\begin{equation}
    L(A, B) = - \sum_{(i,j)} \frac{1}{2} 
    \left[ \log \left( \frac{e^{\langle h^{A}_i, h^{B}_j \rangle / \tau}}
    {\sum_k e^{\langle h^{A}_i, h^{B}_k \rangle / \tau}} \right) 
    + \log \left( \frac{e^{\langle h^{A}_i, h^{B}_j \rangle / \tau}}
    {\sum_k e^{\langle h^{A}_k, h^{B}_j \rangle / \tau}} \right) 
    \right]
    \label{eq:infoNCE}
\end{equation}
where $\langle \cdot, \cdot \rangle$ is any distance function, $A, B \in \{C, R, X\}$, and $\tau$ is the temperature that shapes the distribution. This objective encourages embeddings of semantically corresponding instances to remain proximal in the latent space while distant from others, ensuring that the CXR encoder learns to properly connect the latent representations of CT and the corresponding textual report derived by CT-CLIP. Finally, we define the learning objective of X2CT-CLIP below to incorporate contrastive loss across different pairs:
\begin{equation}
    L_{\text{X2CT}}(C,R,X) = \alpha L(C, R) + \beta L(X, R) + \gamma L(X, C)
    \label{eq:final_loss}
\end{equation}
with different weighting factors $\alpha$, $\beta$, and $\gamma$. We set $\beta = \gamma = 1$, and $\alpha = 0$ in Eq.~\ref{eq:final_loss} (as demonstrated in Fig~\ref{fig:method}-Left), freezing the parameters of CT-CLIP and training only the CXR encoder $f_X(\cdot)$ to preserve the latent space structure and retain the knowledge embedded in CT-CLIP. This has the added advantage of significantly lowering the demand for computation resources during CXR encoder pretraining. 

\subsection{Implementation Details}
We follow the same data preprocessing pipeline as in CT-CLIP \cite{ctclip} for the CT volumes before projecting them to CXR images. The image input size and the latent feature dimension of the CXR encoder are set to 224 and 512, respectively. Finally, we train the CXR encoder for 50 epochs using the proposed objective function (Eq~\ref{eq:final_loss}) with learning rate $5e^{-5}$, batch size $360$, $\tau=0.07$, and optimized using AdamW \cite{adamw} in the Pytorch framework \cite{pytorch}. Our tri-modal CL framework is agnostic to the CXR model architecture, and all pretraining and experiments are conducted on a single 40GB NVIDIA A100 GPU.

\section{Experiments and Results}
\begingroup  
    \setlength{\tabcolsep}{3pt}  
    \setlength\dashlinedash{0.3pt}     
    \setlength\dashlinegap{1.5pt}      
    \setlength\arrayrulewidth{0.3pt}   
    \begin{table}[t!]
        \centering
        \caption{Multi-label datasets. ZS: zero-shot. FS: few-shot adaptation. EV: external validation. CT-R: CT reports. $^1$Simulated CXR. $^2$Custom curated from MIMIC\cite{mimic-cxr}. $^\dagger$This is the validation split of CT-RATE, as no test split is available, but for naming consistency with other datasets, we refer to it herein as the \texttt{Test} subset. }
        \begin{tabular}{c:c:c:c:c}
            \thickhline
            Dataset & Modalities & Task & Train & Test\\
            \hline
            CT-RATE \cite{ctclip} & $\text{CXR}^1$/CT/CT-R & Pretrain/Retrieval/ZS/FS & 50,188 & $3038^\dagger$\\
            RadChest-CT \cite{radchest_ct} & $\text{CXR}^1$/CT & Retrieval/EV/ZS/FS & - & 3630\\
            $\text{MIMIC-CT}^2$& CXR/CT-R & Retrieval/EV/ZS & - & 256\\
            \thickhline
        \end{tabular}
        \label{tab:dataset}
    \end{table}
\endgroup  
\subsection{Multi-Label Datasets for Validations}
We summarize the three multi-label datasets used for pretraining and validation in Table~\ref{tab:dataset}. CT-RATE with simulated CXRs was created as noted above. For RadChest-CT \cite{radchest_ct}, we follow the same procedure described in Sec~\ref{sec:synxray} to simulate CXR images from CT scans. To validate our model on real CXR images with corresponding CT reports, we curated a subset of CXRs from MIMIC and MIMIC-CXR \cite{mimic-cxr} matched by hadm\_id with their corresponding discharge notes and radiology reports, referred to as MIMIC-CT; We then followed \cite{llmLabeler} to extract CT labels from these reports using LLaMA-8B-Instruct~\cite{dubey2024llama}.

We performed top-k cross-modal retrieval, zero-shot (ZS), and few-shot (FS) multi-label prediction tasks. All validations were conducted on the \texttt{Test} splits. For the FS adaptation task in CT-RATE, we sampled from the \texttt{Train} split and evaluated on its \texttt{Test} subset. For RadChest-CT, we focused on CT scan labels while leaving the remaining labels for future work. We then sampled a subset from the \texttt{Test} split to fine-tune the classifier and validated the remaining instances for the FS adaptation task. MIMIC-CT is excluded from FS adaptation due to its limited size, which is insufficient for the FS multi-label prediction task.

We also examined the generalizability of learned FS classifiers, with linear probing trained on CT-RATE, through external validation on RadChest-CT and MIMIC-CT. We identified overlapping labels between CT-RATE and RadChest-CT, as well as between CT-RATE and MIMIC-CT, for evaluation. We then sampled instances from \texttt{Train} split of CT-RATE to fine-tune the linear classifier and evaluate its performance on the \texttt{Test} split of MIMIC-CT and RadChest-CT. 

Notes: 1) We emphasize that these are multi-label datasets and thus traditional K shots sampling of each label is not feasible. We therefore sampled the data for FS adaptation and external validation tasks following \cite{stratification}; and 2) CT-CLIP results are provided for reference where applicable, as they are performed on direct CT volume (or CT report) queries as opposed to CXRs.
\subsection{Experiments Overview}
Our approach is extensively validated against CXR-based foundation models \cite{gloria, medclip, cxrclip}, which utilize ResNet \cite{resnet}, Swin transformer\cite{swin}, and DenseNet \cite{densenet} as vision backbones. We also include comparisons with BI-Mamba \cite{bimamba}, a model specifically trained to identify CT-level pathologies using simulated CXRs. We evaluated ZS, FS, and external validation on multi-label classification tasks using the Area Under the Receiver Operating Characteristic (AUC) and Precision-Recall AUC (PR) metrics. The statistical significance of AUC differences was determined using the two-tailed DeLong test \cite{delong} computed using \cite{fast_delong} at $\alpha=0.05$. We used top-k recall ($\text{R}_k$) to evaluate cross-modal retrieval performance.
\begingroup  
    \setlength{\tabcolsep}{2pt}  
    \setlength\dashlinedash{0.3pt}     
    \setlength\dashlinegap{1.5pt}      
    \setlength\arrayrulewidth{0.3pt}   
    
    \begin{table}[t!]
        \centering
        \caption{Cross-modal retrieval from CXR. CT-V Retri: CT volume retrieval. CT-R retri: CT report retrieval. Note that CT-CLIP is provided for reference only, as it utilizes either the CT report or CT volume, rather than a CXR as query.}
        \begin{tabular}{cc:cc:cc:cc:cc}
            \thickhline
            \multirow{3}{*}{Method} & \multirow{3}{*}{Backbone} & \multicolumn{4}{c:}{\textbf{CT-RATE}} & \multicolumn{2}{c:}{\textbf{RadChest-CT}} & \multicolumn{2}{c}{\textbf{MIMIC-CT}}\\
            & & \multicolumn{2}{c:}{CT-V Retri} & \multicolumn{2}{c:}{CT-R Retri} & \multicolumn{2}{c:}{CT-V Retri}& \multicolumn{2}{c}{CT-R Retri}\\
            & & $\text{R}_5\uparrow$ & $\text{R}_{10}\uparrow$ & $\text{R}_5\uparrow$ & $\text{R}_{10}\uparrow$ & $\text{R}_5\uparrow$ & $\text{R}_{10}\uparrow$ & $\text{R}_5\uparrow$ & $\text{R}_{10}\uparrow$\\
            \hline
            \multirow{2}{*}{MedCLIP} & ResNet & 0.001 & 0.001 & 0.001 & 0.001 & 0.001 & 0.001 & 0.012  & 0.023\\
                                      & Swin & 0.001 & 0.002 & 0.001 & 0.002 & 0.001 & 0.001 & 0.012  & 0.023\\
            \hdashline[1pt/2pt]
            \multirow{2}{*}{CXR-CLIP} & ResNet & 0.000 & 0.001 & 0.001 & 0.002 & 0.001 & 0.001 & 0.004  & 0.004\\
                                      & Swin & 0.000 & 0.001 & 0.000 & 0.001 & 0.001 & 0.001 & 0.019  & 0.019\\
            \hdashline[1pt/2pt]
            CT-CLIP & 3DViT & 0.030 & 0.055 & 0.029 & 0.051 & - & - & - & -\\
            \hline
            \multirow{2}{*}{Ours} & ResNet & \underline{0.118} & \underline{0.162} & \textbf{0.048} & \underline{0.077} & \underline{0.043} & \underline{0.062}  & \textbf{0.113}  & \underline{0.141}\\
                                  & Swin & \textbf{0.129} & \textbf{0.181} & \underline{0.047} & \textbf{0.078} & \textbf{0.045} & \textbf{0.063} & \textbf{0.113}  & \textbf{0.160}\\
            \thickhline
        \end{tabular}
        \label{tab:ct_merged_retrieval}
    \end{table}
\endgroup  

\subsection{Top-K Cross-modal Retrieval Task}
Table~\ref{tab:ct_merged_retrieval} presents the results of CT volume and CT report retrieval queried by CXR across all three datasets. Surprisingly, models pretrained with our X2CT-CLIP consistently outperform the CT-CLIP teacher model in the recall metric, a potential indicator that the CXR encoder learned a better latent space than the originally pre-aligned CT encoder. We hypothesize that the push-and-pull property in Eq.~\ref{eq:infoNCE} helps CXR embeddings find better locations in CT-CLIP latent space that align more effectively with CT volume and CT report features. Furthermore, CXR-based foundation models struggled in the $\text{R}_k$ metric for cross-modal retrieval. Therefore, our knowledge transfer mechanism demonstrates superior tri-modal alignment in latent space compared to existing baselines. 
\subsection{Multi-Label Classification Tasks in CT using CXRs}
\begingroup  
    \setlength{\tabcolsep}{5pt}  
    \setlength\dashlinedash{0.3pt}     
    \setlength\dashlinegap{1.5pt}      
    \setlength\arrayrulewidth{0.3pt}   
    \begin{table}[t!]
        \centering
        \caption{Zero-shot multi-label classification. Note, CT-CLIP is queried by CT volume and for reference only. $^*$The best result that has a significant ($p < 0.05$) difference in AUC compared to the closest baseline, based on DeLong's two-tailed test \cite{delong}.}
        \begin{tabular}{cc:cc:cc:cc}
            \thickhline
            \multirow{2}{*}{Method} & \multirow{2}{*}{Backbone} & \multicolumn{2}{c:}{\textbf{CT-RATE}} & \multicolumn{2}{c:}{\textbf{RadChest-CT}} & \multicolumn{2}{c}{\textbf{MIMIC-CT}}\\
            & & AUC$\uparrow$ & PR$\uparrow$ & AUC$\uparrow$ & PR$\uparrow$ & AUC$\uparrow$ & PR$\uparrow$ \\
            \hline
            \multirow{2}{*}{MedCLIP} & ResNet & 0.378 & 0.220 & 0.468 & 0.445 & 0.456 & 0.346 \\
                                     & Swin & 0.518 & 0.296 &  0.514 & 0.476 & 0.540 & 0.419\\
            \hdashline[1pt/2pt]
            \multirow{2}{*}{CXR-CLIP} & ResNet & 0.448 & 0.242 & 0.469 & 0.442 & 0.475 & 0.387\\
                                     & Swin & 0.516 & 0.286 & 0.525 & 0.477 & 0.462 & 0.361\\
            \hdashline[1pt/2pt]
            CT-CLIP  & 3DViT & 0.697 & 0.413 &  0.617 & 0.536 & - & -\\
            \hline
            \multirow{2}{*}{Ours} & ResNet & \textbf{0.716$^*$} & \textbf{0.438} & \textbf{0.645$^*$} & \textbf{0.550} & \textbf{0.567$^*$} & \textbf{0.430}\\
                                  & Swin & \underline{0.714} & \underline{0.435} &  \underline{0.644} & \underline{0.548} & \underline{0.557} & \underline{0.424} \\
            \thickhline
        \end{tabular}
        \label{tab:zero_shot}
    \end{table}
\endgroup  

\textbf{Zero-shot evaluation}: We show the ZS performance across the three datasets in Table~\ref{tab:zero_shot}. Baselines with incompatible embedding sizes with the CT-CLIP text encoder are omitted. Backbones pretrained with X2CT-CLIP demonstrated improved tri-modal alignment in latent space compared to CT-CLIP and other CXR foundation models. This improvement remained across simulated and real CXR inputs. 
\begingroup  
    \setlength\dashlinedash{0.3pt}     
    \setlength\dashlinegap{1.5pt}      
    \setlength\arrayrulewidth{0.3pt}   
    \begin{table}[t!]
        \centering
        \caption{Multi-Label classification in CT. \textbf{(a) FS adaptation via linear probing}: RadChest-CT and CT-RATE using classical FS. \textbf{(b) External validation}: Linear probing trained on the CT-RATE (the pretraining dataset), and evaluated on RadChest-CT or MIMIC-CT. $^*$The best result with a significant ($p < 0.05$) difference in AUC compared to the closest baseline, based on DeLong's two-tailed test \cite{delong}.}
        \begin{tabular}{cc:cc:cc:cc:cc}  
            \multicolumn{10}{l}{\textbf{(a) Few-shot (FS) adaptation via linear probing}}\\
            \thickhline
            \multirow{3}{*}{Method} & \multirow{3}{*}{Backbone} & \multicolumn{4}{c:}{\textbf{CT-RATE}} & \multicolumn{4}{c}{\textbf{RadChest-CT}} \\
            & & \multicolumn{2}{c:}{FS@20\%} & \multicolumn{2}{c:}{FS@50\%} & \multicolumn{2}{c:}{FS@20\%} & \multicolumn{2}{c}{FS@50\%} \\
            & & AUC$\uparrow$ & PR$\uparrow$ & AUC$\uparrow$ & PR$\uparrow$ & AUC$\uparrow$ & PR$\uparrow$ & AUC$\uparrow$ & PR$\uparrow$ \\  
            \hline
            \multirow{2}{*}{GLoRIA} & ResNet & 0.784 & 0.455 & 0.785 & 0.457 & 0.870 & 0.629 & 0.869 & 0.619\\
                                      & DenseNet & 0.815 & 0.525 & 0.816 & 0.524 & 0.878 & 0.652 & 0.880 & 0.644\\
            \hdashline[1pt/2pt]
            \multirow{2}{*}{MedCLIP} & ResNet & 0.799 & 0.494 & 0.799 & 0.497 & 0.880 & 0.649 & 0.876 & 0.626\\
                                      & Swin & 0.804 & 0.507 & 0.787 & 0.473 & 0.879 & 0.644 & 0.878 & 0.631\\
            \hdashline[1pt/2pt]
            \multirow{2}{*}{CXR-CLIP} & ResNet & 0.826 & 0.549 & 0.828 & 0.551 & 0.878 & 0.670 & 0.881 & 0.669\\
                                      & Swin & 0.828 & 0.549 & 0.831 & 0.553 & 0.879 & 0.671 & 0.883 & 0.669\\  
            \hdashline[1pt/2pt]
            Multi-View & BI-Mamba & 0.772 & 0.43 & 0.779 & 0.448 & 0.871 & 0.641 & 0.869 & 0.633\\
            \hline
            \multirow{2}{*}{Ours} & ResNet & \underline{0.840} & \underline{0.565} & \underline{0.843} & \underline{0.577} & \underline{0.887} & \textbf{0.683} & \underline{0.893} & \underline{0.687}\\
                                  & Swin & $\textbf{0.841}^*$ & \textbf{0.566} & $\textbf{0.847}^*$ & \textbf{0.579} & $\textbf{0.888}^*$ & \underline{0.681} & $\textbf{0.894}^*$ & \textbf{0.692} \\
            \thickhline
            \\
            \multicolumn{10}{l}{\textbf{(b) External validation (classifier trained on CT-RATE)}}\\
            \thickhline
            \multirow{3}{*}{Method} & \multirow{3}{*}{Backbone} & \multicolumn{4}{c:}{\textbf{RadChest-CT}} & \multicolumn{4}{c}{\textbf{MIMIC-CT}}\\
            & & \multicolumn{2}{c:}{FS@5$\%$} & \multicolumn{2}{c:}{FS@10$\%$} & \multicolumn{2}{c:}{FS@5$\%$} & \multicolumn{2}{c}{FS@10$\%$}\\
            & & AUC$\uparrow$ & PR$\uparrow$ & AUC$\uparrow$ & PR$\uparrow$ & AUC$\uparrow$ & PR$\uparrow$ & AUC$\uparrow$ & PR$\uparrow$ \\  
            \hline
            \multirow{2}{*}{GLoRIA} & ResNet & 0.694 & 0.432 & 0.692 & 0.431 & 0.709 & 0.313 & 0.726 & 0.339\\
                                      & DenseNet & 0.712 & 0.496 & 0.719 & 0.497 & 0.692 & 0.316 & 0.710 & 0.330\\
            \hdashline[1pt/2pt]
            \multirow{2}{*}{MedCLIP} & ResNet & 0.684 & 0.441 & 0.689 & 0.453 & 0.754 & 0.365 & 0.770 & 0.388\\
                                      & Swin & 0.700 & 0.457 & 0.701 & 0.455 & 0.752 & 0.383 & 0.764 & 0.408\\
            \hdashline[1pt/2pt]
            \multirow{2}{*}{CXR-CLIP} & ResNet & 0.721 & 0.512 & 0.721 & 0.505 & 0.684 & 0.314 & 0.688 & 0.304\\
                                      & Swin & 0.722 & 0.513 & 0.719 & 0.512 & 0.639 & 0.242 & 0.665 & 0.272\\  
            \hdashline[1pt/2pt]
            Multi-View & BI-Mamba & 0.663 & 0.412 & 0.666 & 0.424 & 0.742 & 0.360 & 0.733 & 0.369\\
            \hline
            \multirow{2}{*}{Ours} & ResNet & $\textbf{0.735}^*$ & $\textbf{0.559}$ & \underline{0.728} & \underline{0.555} & $\textbf{0.790}^*$ & \textbf{0.434} & \underline{0.766} & \underline{0.412}\\
                                  & Swin & \underline{0.731} & \underline{0.555} & $\textbf{0.733}^*$ & \textbf{0.558} & \underline{0.760} & \underline{0.412} & $\textbf{0.794}^*$ & \textbf{0.438}\\
            \thickhline
        \end{tabular}
        \label{tab:abnormality_detection}
    \end{table}
\endgroup  

\textbf{Few-shot adaptation via linear probing}: We performed FS adaptation on RadChest-CT and CT-RATE for multi-label classification by linear probing on 20\% and 50\% of each dataset. As shown in Table~\ref{tab:abnormality_detection}a, our method consistently outperformed BI-Mamba and CXR foundation models across all settings, achieving the highest AUC ($p<0.05$) and PR score in both CT-RATE and RadChest-CT. These results illustrate the efficacy of our CT-to-CXR knowledge transfer strategy in enabling robust CT-level disease prediction using limited CXR data.

\textbf{External validation}: Different from the traditional FS setting and to measure the robustness of X2CT-CLIP, we also performed external validation on our pretrained CXR encoder under a highly size constrained data setting by fine-tuning the classifier with 5\% and 10\% of \texttt{Train} split in CT-RATE and inference on other datasets. As shown in Table~\ref{tab:abnormality_detection}b, our pretrained models showed better overall performance by consistently achieving higher AUC scores ($p < 0.05$) than other models in both simulated and real CXR settings. The larger margin of improvement in the PR metric also suggests that our learning strategy may effectively reduced false positives and false negative predictions, and thus identify high-risk patients while keeping false alarms manageable in multi-abnormality detection.

\textbf{Ablation study on the learning objective}: We analyzed the impact of including CT reports ($\beta$) and CT volumes ($\gamma$) in our loss function (Eq.~\ref{eq:final_loss}) by evaluating on the \texttt{Test} split of CT-RATE. Table~\ref{tab:ablation} shows that while removing textual or CT volume knowledge may benefit their respective tasks in top-k recall metric, it results in an approximate loss of $1.5\%$ in AUC and $2\%$ in PR for the multi-label detection task compared to our proposed objective (last row). This underscores the need to integrate both modalities for robust recognition of multi-abnormality in CT scans using CXR.
\begingroup  
    \setlength{\tabcolsep}{3pt}  
    \setlength\dashlinedash{0.3pt}     
    \setlength\dashlinegap{1.5pt}      
    \setlength\arrayrulewidth{0.3pt}   
    \begin{table}[t!]
        \centering
        \caption{Ablation study based on CT-RATE data. FT: fine-tuning classifier. Validations are conducted on the \texttt{Test} split of CT-RATE.}
        \begin{tabular}{c:cc:cc:ccc:ccc}
            \thickhline
            \multirow{2}{*}{Backbone} & \multirow{2}{*}{$\beta$} & \multirow{2}{*}{$\gamma$} & \multicolumn{2}{c:}{FT} & \multicolumn{3}{c:}{CT Report Retrieval} & \multicolumn{3}{c}{CT Volume Retrieval} \\
            &  &  & AUC$\uparrow$ & PR$\uparrow$ & $\text{R}_{5}\uparrow$ & $\text{R}_{10}\uparrow$ & $\text{R}_{50}\uparrow$ & $\text{R}_{5}\uparrow$ & $\text{R}_{10}\uparrow$ & $\text{R}_{50}\uparrow$\\
            \hline
            \multirow{3}{*}{ResNet} & 1 & 0 & 0.833 & \underline{0.560} & \underline{0.033} & \underline{0.056} & \underline{0.160} & 0.035 & 0.053 & 0.169\\
                                    & 0 & 1 & \underline{0.835} & 0.555 & 0.022 & 0.035 & 0.124 & \textbf{0.163} & \textbf{0.228} & \textbf{0.493}\\
                                    & 1 & 1 & \textbf{0.847} & \textbf{0.582} & \textbf{0.048} & \textbf{0.077} & \textbf{0.193} & \underline{0.118} & \underline{0.162} & \underline{0.401}\\
            \thickhline
        \end{tabular}
        \label{tab:ablation}
    \end{table}

\endgroup  

\section{Conclusion}
In this study, we addressed the challenge of predicting multi-abnormality in CT from CXR by proposing X2CT-CLIP, the first tri-modal contrastive learning framework that bridges the modality gap between CXR and CT. By aligning CXR to the pre-aligned CT and CT report representations, our method outperformed state-of-the-art models in all validation tasks while requiring much lower hardware resources in model training. These results demonstrated the feasibility of using CXR for CT-level disease prediction, offering a scalable and efficient alternative for clinical screening, particularly in a restricted data regime.

\bibliographystyle{splncs04}
\bibliography{reference}
\end{document}